# LegalPro-BERT: Classification of Legal Provisions by fine-tuning BERT large language model

**Amit Tewari**

Email: atewari35@gatech.edu

ORCID: https://orcid.org/0000-0002-6821-3833

**Abstract**

A contract is a type of legal document commonly used in organizations. Contract review is an integral and repetitive process to avoid business risk and liability. Contract analysis requires the identification and classification of key provisions and paragraphs within an agreement. Identification and validation of contract clauses can be a time-consuming and challenging task demanding the services of trained and expensive lawyers, paralegals or other legal assistants. Classification of legal provisions in contracts using artificial intelligence and natural language processing is complex due to the requirement of domain-specialized legal language for model training and the scarcity of sufficient labeled data in the legal domain. Using general-purpose models is not effective in this context due to the use of specialized legal vocabulary in contracts which may not be recognized by a general model. To address this problem, we propose the use of a pre-trained large language model which is subsequently calibrated on legal taxonomy. We propose LegalPro-BERT, a BERT transformer architecture model that we fine-tune to efficiently handle classification task for legal provisions. We conducted experiments to measure and compare metrics with current benchmark results. We found that LegalPro-BERT outperforms the previous benchmark used for comparison in this research.

# 1 Introduction

A contract is a legally binding agreement between two or more parties that stipulates the terms of the agreement. Modern organizations maintain a large inventory of contracts and other legal documents. A typical large Fortune organization manages up-to 40,000 active contracts at any point in time (Pery and Simon 2019). A substantial amount of time and cost is involved in manually reviewing and analyzing these contracts for due diligence and risk analysis. According to a study (Dodd 2018), more than 50% of contract review tasks e.g. clause identification and classification (Zhao 2019) are repetitive in nature requiring legal domain knowledge and understanding.

However, due to the extremely large scale and ever-increasing volume of contracts that an organization must deal with, manually segregating paragraphs in these contracts to derive any actionable information is a monumental task for any law firm or legal department. Therefore automatic categorization and classification of legal clauses and provisions using natural language processing and artificial intelligence has been of great academic as well as commercial interest (Katz et al. 2023, DALE 2019).

The primary research interest for this paper is the classification of provisions found in contracts into topics. This involves two important considerations: A) Sophisticated neural network classification methods require a large set of labeled curated data which can be a difficult and costly exercise, B) The classifier models trained on general language corpus are not efficient for the legal classification task, because legal texts have a specialized language and distinctive vocabulary with peculiar expressions and terms that do not commonly occur in day-to-day spoken or written language.

The focus of this research is the usage of transfer learning (Zhuang et al. 2019) techniques because they look promising to address the two challenges above. The central tenet in these techniques is to first train a general-purpose language model to serve as a base model by training on a very large general corpus but without using any task-specific labeled dataset. At this step, the base model only encodes the semantic information as lower-order features. Then, fine-tune (Dodge et al. 2020) and customize this foundational model for the classification task using relatively fewer labeled data points and weights learned in the base model. By following this approach, higher-order feature representations are fine-tuned making them more efficient at the specific task by utilizing the lower-order general features learned in the original base model without having to explicitly re-create them. This avoids the need to re-train a large model using a large dataset from the beginning for the specific task.

The aim of this research is to examine the applicability of fine-tuning a pre-trained large language model for legal domain classification task. In support of that, the main topic of provision in a contract will be predicted. The corpus of contracts used is obtained from public domain filings in the Securities and Exchange Commission (SEC) curated in Legal General Language Understanding Evaluation (LexGLUE) (Chalkidis et al. 2021) which uses LEDGAR dataset (Tuggener et al. 2020). Each label in the dataset represents the single most important topic (subject) in the corresponding contract provision (refer Table 1 for sample data format). The provisions and their count are given in appendix Table A.

**Table 1** Sample data representation

| Clause | Label | Category |
|---|---|---|
| Except as otherwise set forth in this Debenture, the Company, for itself and its legal representatives, successors and assigns, expressly waives presentment, protest, demand, notice of dishonor, notice of nonpayment, notice of maturity, notice of protest, presentment for the purpose of accelerating maturity, and diligence in collection. | 97 | Waivers |
| No ERISA Event has occurred or is reasonably expected to occur that, when taken together with all other such ERISA Events for which liability is reasonably expected to occur, could reasonably be expected to result in a Material Adverse Effect. Neither Borrower nor any ERISA Affiliate maintains or contributes to or has any obligation to maintain or contribute to any Multiemployer Plan or Plan, nor otherwise has any liability under Title IV of ERISA. | 39 | Erisa |
| This Amendment may be executed by one or more of the parties hereto on any number of separate counterparts, and all of said counterparts taken together shall be deemed to constitute one and the same instrument. This Amendment may be delivered by facsimile or other electronic transmission of the relevant signature pages hereof. | 26 | Counterparts |
| From time to time, as and when required by the Surviving Corporation or by its successors or assigns, there shall be executed and delivered on behalf of Ashford (DE) such deeds and other instruments, and there shall be taken or caused to be taken by it all such further and other action, as shall be appropriate, advisable or necessary in order to vest, perfect or confirm, of record or otherwise, in the Surviving Corporation the title to and possession of all property, interests, assets, right... | 45 | Further Assurances |
| Commencing March 7, 2016 and during the Employment Period, the Company shall pay to the Executive a base salary at the rate of no less than $750,000 per calendar year (the "Base Salary"), less applicable deductions, and prorated for any partial month or year, as applicable. The Base Salary shall be reviewed for increase by the Compensation Committees of AFG and AAC (the "Compensation Committees") no less frequently than annually and may be increased in the discretion of the Compensation Comm... | 11 | Base Salary |

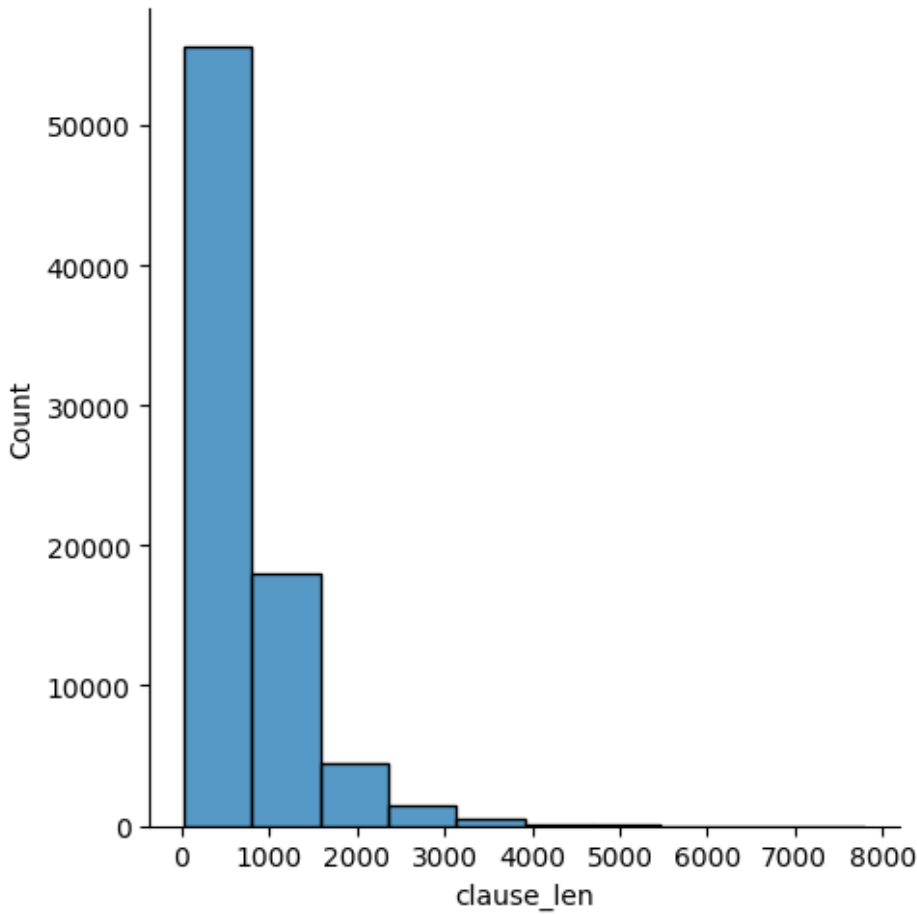

**Fig. 1 Clause length distribution**

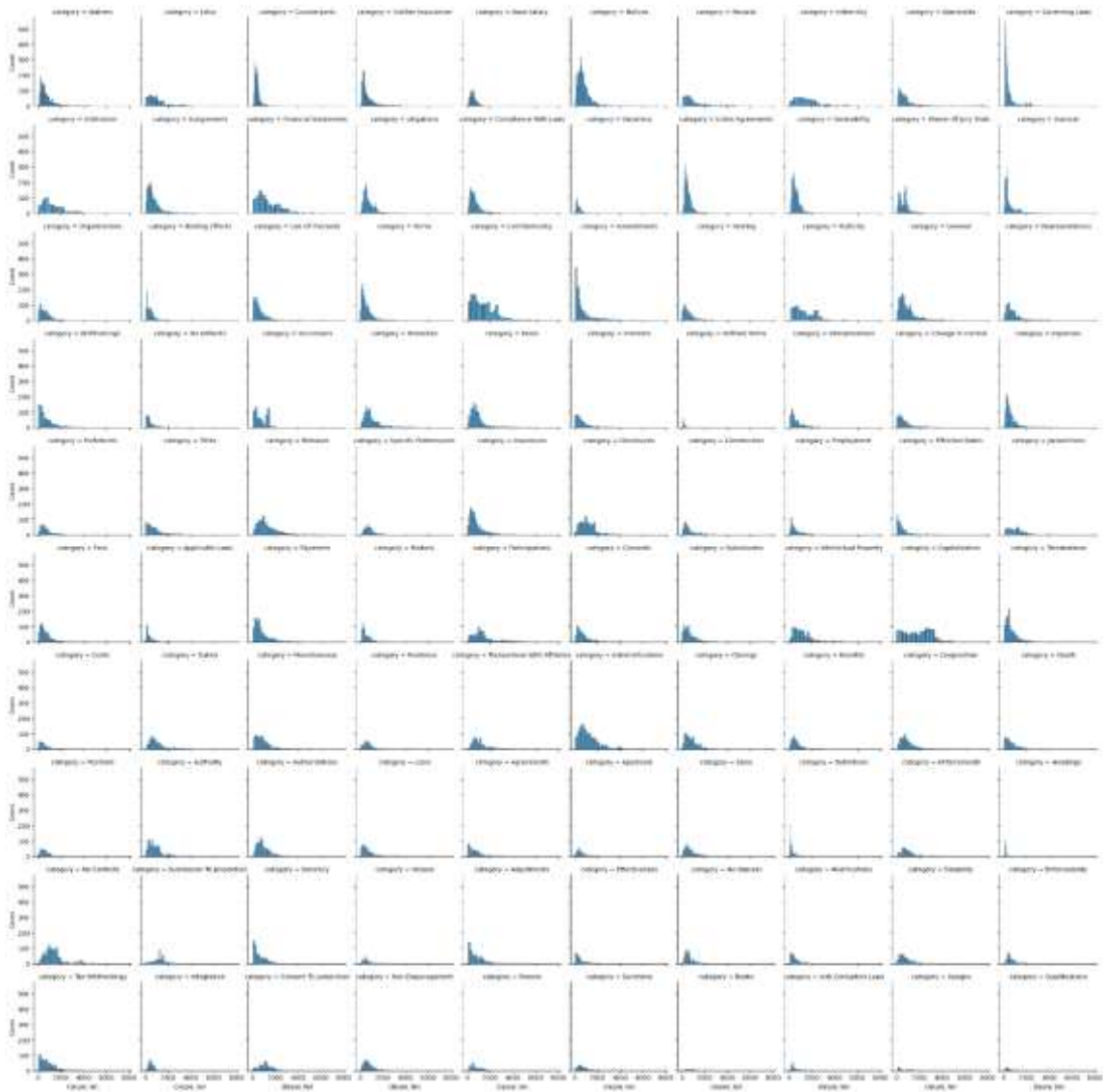

**Fig. 2 Distribution of Clause lengths with each class**

Pre-trained transformer models such as BERT (Devlin et al. 2019), GPT (OpenAI 2023) represent the current state of the art in many natural language processing (NLP) tasks.

With this in mind, we propose fine-tuning BERT model (Sun et al. 2019) to automate classification of legal provisions into pre-defined legal taxonomical categories.

The main contributions of this research are as follows:

- We present LegalPro-BERT, a large language model based on BERT-large (Devlin et al. 2019) for legal provision classification task.
- We compare LegalPro-BERT using benchmark established in LexGLUE LEDGAR classification dataset (Chalkidis et al. 2021).
- We achieve better performance than LexGLUE classification benchmark on LEDGAR data.
- We conduct experiments to examine several facets of the model, including effects of data pre-processing techniques on model accuracy, fine-tuning only a small subset of model layers thereby decreasing overall time and improvement in prediction by adjusting the pre-trained features to the new data.

The remainder of the paper is structured in the following manner: First, relevant literature in legal domain classification tasks using neural network models is discussed (Section 2). Then, the model training steps leading to the final model used in this paper is explained (Section 3). In the next section, structure of the experiment in the current study is explained (Section 4). Then we present the results of our experiment on the legal clause classification dataset (Section 5). We further examine LegalPro-BERT from different contexts and provide analysis of our experiment in Section 6. Next, we go on to enumerate limitations (Section 7) and propose future work (Section 8). We then conclude in Section 9.

## 2 Related Work

In this section we present an overview of different legal text classification studies done using machine learning and deep learning. One of the earliest attempts for classification of legal documents was done involved statute documents (Curran and Thompson 1997) using distribution of frequencies in the documents and training a decision tree classifier. This early study had an f-score of 0.56. Another study (de Maat and Winkels 2010) proposed classification of sentences using sentence structure. The study centered around sentences in Dutch Law over a relatively small set of 530 sentences only. Purpura and Hillard (2006) experimented with classification of Congressional legislations into 226 topic areas using Support Vector Machines (Zhang 2012). The overall accuracy was 71% across over 200 possible categories. Ruggeri et al. (2022) propose a framework for classification of 'unfair clause' in Terms of service using memory-augmented neural networks and produced f1-score between 0.52-0.66 for various categories of the 2346 unfair clauses in a corpus of 21,063 sentences. Zhao (2019) proposed an expert based system method for classification of contractual terms and termination in 3 categories - conditions, warranties and intermediate terms. Braun and Matthes (2022) compared different machine learning approaches for classifying 6000 German and English Terms and Conditions clauses across 37 subtopics. The best model in this study was identified as BERT with an f1-score of 0.91. Braun and Matthes (2021) focus on finding unfavorable and void clauses in Terms and Conditions from German online shopping by using a fine-tuned BERT model. They used a dataset of 1,186 clauses across 22 categories. They only report accuracy of 0.9 which does not takes into account any class imbalance across the clause topics. Aggarwal et al. (2021) proposed predicting 5 types of clauses that may be relevant to a contract and produce relevant content for automatic contract authoring basis the contract context. Indukuri and Krishna (2010) presented an approach to extract and classify payment and non-payment related clauses from e-contract documents using a 2-step combination of human-in-loop and binary support vector machine (SVM). Shah et al. (2018) proposed legal clause extraction in a corpus of 100 pre-tagged contracts using paragraph boundary detection and segmentation.

Above studies used a relatively small corpus ranging from 100 to few thousand data points only. Also, there are few studies to evaluate application of fine-tuned pre- trained model for legal clause classification and categorization. In our opinion, the closest work similar to our research was done by Chalkidis et al. (2021) proposed a benchmark dataset for legal language understanding using LEDGAR dataset for legal clause classification. However, to the best of our knowledge there were no subsequent studies done to either improve the accuracy achieved in the Chalkidis et al. (2021) paper or employ fine-tuning large models for legal clause classification over a large set of contracts and constituent paragraphs.

# 3 Method

In this section we present an introduction to relevant neural networks followed by implementation of our fine-tuned BERT model for legal clause classification named as LegalPro-BERT.

## 3.1 Preliminaries

### 3.1.1 RNN

A recurrent neural network (RNN) (Rumelhart et al. 1986) is a type of artificial neural network (Uhrig 1995) that incorporates looping connections between nodes, enabling the output of a node to influence input to the same node. As a result, RNNs demonstrate temporal dynamic behavior which implies that any future input can potentially cause the system to alter the output. Rooted in feed-forward neural networks, RNNs leverage their internal past state (memory) to effectively handle input sequences of varying lengths, making them suitable for temporal problems like recognition of speech and language translation. This capability of RNNs allows them to process arbitrary sequences of inputs. RNNs start to show in-efficiencies in long sequences when the context is located at a further distance from the point where it is applicable.

### 3.1.2 LSTM

The Long Short-Term Memory (LSTM) (Hochreiter and Schmidhuber 1997) network belongs to the category of recurrent neural networks (RNNs) and was specifically designed to address the issue of vanishing gradients (Hochreiter 1998) encountered in conventional RNNs. Unlike traditional RNNs, LSTM demonstrates resilience to window length, offering a short-term memory that can extend over thousands of time steps. This characteristic has earned it the name 'long short-term memory.' The key to LSTM's ability to capture long-term dependencies within a sequence lies in its utilization of 'forget' and 'update' cells, enabling the network to maintain and update relevant information effectively. LSTMs also suffer from the same issues as those in RNNs namely sequential computations and the inability to determine the context in case of long sequences.

### 3.1.3 Transformer

The Transformer (Vaswani et al. 2023) architecture due to its capability to simultaneously process input sequences, is widely used in large language models for language understanding tasks. Transformer contextualizes input tokens using a self-attention mechanism at each layer by modeling associations between all words in the input with- out consideration of order. It offers rapid training in comparison to recurrent neural networks. The encoder part of the Transformer comprises numerous identical layers of self-attention and feed-forward networks. By enabling parallelization, the Trans- former overcomes the limitations of RNN-based approaches, making it capable to process information across larger sequences. Development of pre-trained systems like BERT (Devlin et al. 2019) has been possible due to advancements proposed through Transformers.

### 3.1.4 BERT

Bidirectional Encoder Representations from Transformers, or BERT exploits bidirectional training to Transformer for language modelling tasks. This is key differentiator to previous architectures which processed a sequence from a single direction only at a time. A bidirectional language model has a more thorough understanding of language context compared to single-direction language model. There are many advantages in using an already pre-trained large language model since it reduces computational requirement and thereby cost and starting with a base state-of-the-art model without having to train from ground-up. BERT is already pre-trained to understand syntax as well as language semantic thus it can be fine-tuned using a relatively small legal corpus of labelled clauses for specific task such as classification of legal provisions.

## 3.2 LegalPro-BERT for Classification of Legal Provisions by fine-tuning BERT

In this subsection we explain BERT implementation as done in current research: 1) fine-tuning on legal corpus 2) classification using BERT, 3) text pre-processing pipeline.

### 3.2.1 Fine-tuning on legal domain corpus

We fine-tune the pre-trained BERT model using supervised transfer learning approach to re-train its weights on a legal corpus. Fine-tuning was done only on a subset of network layers and rest of the layers were fixed and inherited from base model.

### 3.2.2 LegalPro-BERT for text classification

BERT pre-trained model already contains high level language features extracted from large text data like Toronto BookCorpus (Zhu et al. 2015) (800M words) and English Wikipedia (2,500M words). Text classification is achieved by adding a dense layer after the last hidden state on top of the Transformer output [CLS] token. After loading the pre-trained BERT model and corresponding tokenizer, we pre-process the input data and generate tokens. After that the classifier neural network is trained in supervised manner by passing the labeled classification dataset. We used BERT-large uncased model with 24-layers, 1024-hidden, 16-attention-heads, 340M parameters with model summary presented in Table 2 below.

**Table 2** Model Summary

| Layer (type) | Output Shape | Param # | Connected to |
|---|---|---|---|
| text (InputLayer) | [(None,)] | 0 | [] |
| keras_layer (KerasLayer) | 'input_mask': (None, 128),'input_word_ids':(None, 128),'input_type_ids': (None, 128) | 0 | ['text[0][0]'] |
| keras_layer_1 (KerasLayer) | 'sequence_output':(None, 128, 1024), 'pooled_output': (None, 1024), 'default': (None, 1024) | 335141889 | ['keras_layer[0][0]', ['keras_layer[0][1]', ['keras_layer[0][2]'] |
| dropout (Dropout) | [(None,1024)] | 0 | ['keras_layer_1[0][1]'] |
| output (Dense) | [(None,100)] | 102500 | ['dropout[0][0]'] |

Total params: 335,244,389
Trainable params: 102,500
Non-trainable params: 335,141,889

### 3.2.3 Text Pre-processing pipeline

We started by converting the text to lower case since we are using BERT-uncased. After that we removed punctuations and stopwords. To help the model further generalize during fine-tuning, we retained only top-100 words in each input text for a given category.

# 4 Experimental Setup

## 4.1 Research Question

We aim to answer the research question - What is the performance of LegalPro-BERT for classification of paragraphs in legal contracts?

## 4.2 Baseline Method

For comparison we use as baseline the benchmark established in LexGLUE study. LexGLUE is a benchmark dataset to evaluate the performance of Natural Language Processing (NLP) methods in legal tasks. Specifically, we compare against LexGLUE multi-label classification of contract paragraphs.

## 4.3 Dataset

In order to fine-tune BERT, we use the LEDGAR contracts dataset from LexGLUE study containing 80K pre-labeled paragraphs spanning more than 9M words and 100 unique classes. Each label represents the single main theme or type of that provision. LEDGAR contains publicly available contracts from the US Securities and Exchange Commission (SEC) filings.

## 4.4 Evaluation Metrics

We use F1-score (Goutte and Gaussier 2005) to measure the overall performance and predictive power of our classification model since F1-score focuses on class-wise performance rather than accuracy. F1 score incorporates both precision and recall into the computation. Further, we use micro-F1($\mu$-F1) and macro-F1(m-F1) to consider class imbalance across the 100 unique classes in our data.

## 4.5 Implementation Details

We utilize the pre-partitioned LEDGAR train, validation and test data containing 60K, 10K, 10K paragraphs respectively. For our implementation, we used a dropout probability of $p = 0.2$. We trained the model for 5 epochs and evaluate on the validation set to choose the best model amongst candidate models. We use Google Colab environment with Tesla V100 GPU and TensorFlow machine learning library.

# 5 Experiment Results

We can see in Table 3 below the results of LegalPro-BERT in comparison to the existing LexGLUE baseline method on legal provision classification task on LEDGAR dataset.

**Table 3** Experimental Results on the LEDGAR dataset

| Method | micro-F1($\mu$-F1) | macro-F1(m-F1) |
| --- | --- | --- |
| LegalPro-BERT (using top-100 most common words) | **0.93** | **0.88** |
| RoBERTa (Large-sized) | 0.88 | 0.83 |
| BERT (Medium-sized) | 0.87 | 0.81 |
| RoBERTa (Medium-sized) | 0.87 | 0.82 |
| DeBERTa | 0.88 | 0.83 |
| Longformer | 0.88 | 0.83 |
| BigBird | 0.87 | 0.82 |
| Legal-BERT (Medium-sized) | 0.88 | 0.83 |
| CaseLaw-BERT (Medium-sized)) | 0.88 | 0.83 |
| BERT-Tiny | 0.83 | 0.74 |
| Mini-LM (v2) | 0.86 | 0.79 |
| Distil-BERT | 0.87 | 0.81 |
| Legal-BERT (Small-sized) | 0.87 | 0.81 |

For both metrics, our LegalPro-BERT model performs better than the models evaluated in LexGLUE LEDGAR classification task on the test dataset consisting of 10,000 samples.

As we notice in Figure 1 below, the rate of loss decay is slow after the second epoch of BERT model fine-tuning. Similarly, rate of improvement observed in accuracy and other metrics plateaus after second epoch. This further demonstrates the efficiency of using a large pre-trained language model as base model and thereafter fine-tuning over a few epochs only.

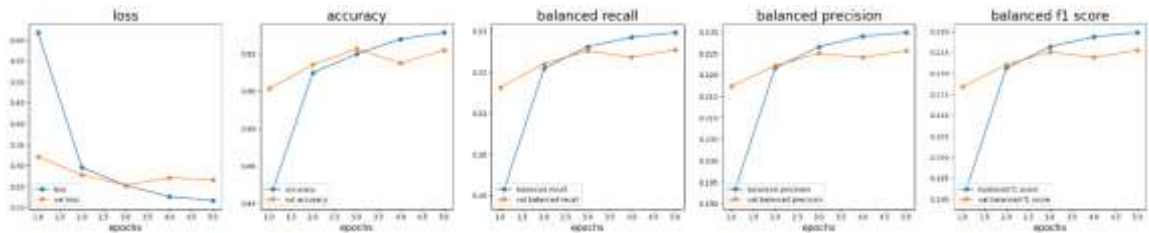

**Fig. 3  Loss**

As we notice in Figure 2 below depicting accuracy in term of actual vs predicted values through the confusion matrix for the test dataset, we observe that the accuracy is high for individual classes as represented by shades of white color. The only exception is class 8 where due to lack of observations in this category in the test dataset, no data is observed in the confusion matrix.

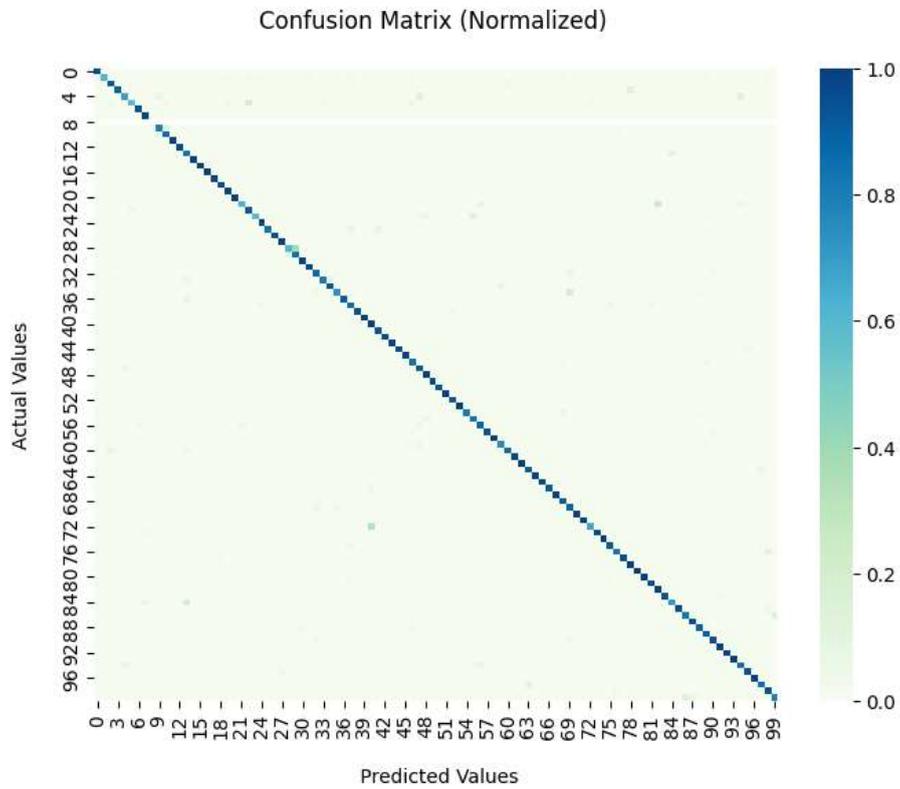

**Fig. 4    Confusion Matrix**

Detailed metrics for individual classes for LegalPro-BERT are presented in appendix Table B.

## 6 Experiment Analysis

This research paper aimed to predict the type or main topic of a legal provision in the LEDGAR dataset. We used pre-segregated clauses from train and validate buckets to fine-train our model and compare the accuracy achieved on test set with that in LexGLUE benchmark . The data from train and validate buckets was used to fine-tune the pre-trained BERT model using legal language embeddings so that the resulting model becomes better in predicting the topic of a clause.

We started by lowercase clause text to train the model and got an f1-score of 0.74.

Next in addition to lower casing we removed punctuations and stopwords and reached f1-score of 0.80.

After that, to further generalize the model during fine-tuning, we retained only top-100 most frequent words in a given topic in each input text. The f1-score of 0.93 achieved by using this most frequent subset approach is major improvement over the 0.88 f1-score in the LexGLUE paper used as benchmark for comparison.

## 7 Limitations

The paper aimed to study and apply the usage of artificial intelligence techniques to automate one specific legal domain problem - clause classification. The data used for the experiment came from LexGLUE LEDGAR benchmark study but there can be other data sources which can be further utilized to examine and evaluate accuracy of fine-tuned BERT models for clause classification. Additionally, the author has presented the paper purely from an academic and scientific background rather than claiming to be an expert in law or legal matters. AI systems are in a state of continuous evolution and require monitoring and professional supervision by legal experts to guard against biases and fallacy.

## 8 Future Work

We have experimented by retaining top 100 most frequent words within a category for each input. Further experimentation using other choices e.g., top 50, 200 etc. most common words in each class can be explored.

There are many other use-cases in legal industry beyond the scope of current research where application of artificial intelligence can be researched e.g., question answering, predicting outcome of legal cases etc.

Additionally, recently announced GPT and LLAMA (Touvron et al. 2023) models have generated considerable interest in the research community for the broad umbrella of use-cases they can handle with little or no re-training. As such, they can be candidates for automation in the current and other legal domain processes.

While the procedure, taxonomy and the models we trained are legal domain specific, the overall approach and idea presented in the study can be easily transferred to other domains e.g. finance, manufacturing, healthcare which use text documents.

# 9 Conclusion

This paper aimed to show application of fine-tuning pre-trained BERT model for clause classification and better understand the performance of transformer model on a specific usecase. LexGLUE LEDGAR dataset is a freely available corpus of 80,000 legal provisions with 100 main topic labels for each provision. The results from this study show that the final model obtained was successfully able to predict the clause labels with 0.93 f1-score for the test set of 10,000 clauses. The BERT-base model was fine-tuned using legal taxonomy for the prediction task so that it could correctly learn how to classify legal clauses.

The results show that a generic model such as BERT can be fine-tuned using legal domain-specific vocabulary to automate tasks which traditionally require legal domain knowledge e.g., to classify legal clauses in a contract, which is generally performed by human and legal experts in legal firms and law departments. After fine-tuning the model can be deployed on-premise during the production stage, so there are no risks associated with data governance and confidentiality since no data is required to be sent to any third-party APIs and servers.

The improvements demonstrated in our analysis over the results in LexGLUE benchmark study shows that artificial intelligence methods can be used to automate tasks in the legal domain.

Overall, this study can not only hasten the progress of research towards the resolution of a critical real-world issue, but also serve as a benchmark for the evaluation of future AI models in other areas besides legal.

# Acknowledgments

The author would like to thank the authors of LEDGAR paper since their corpus of labeled contract legal provisions is used for classification and comparison of benchmark in the current study.

# Appendix A  List of Provisions

**Table A1**: List of Provisions

| Class ID | Provision | Count | Class ID | Provision | Count |
|---|---|---|---|---|---|
| 0 | Adjustments | 462 | 50 | Indemnity | 347 |
| 1 | Agreements | 322 | 51 | Insurances | 1441 |
| 2 | Amendments | 1948 | 52 | Integration | 416 |
| 3 | Anti-Corruption Laws | 155 | 53 | Intellectual Property | 607 |
| 4 | Applicable Laws | 474 | 54 | Interests | 459 |
| 5 | Approvals | 230 | 55 | Interpretations | 541 |
| 6 | Arbitration | 500 | 56 | Jurisdictions | 264 |
| 7 | Assignments | 1730 | 57 | Liens | 417 |
| 8 | Assigns | 38 | 58 | Litigations | 1329 |
| 9 | Authority | 680 | 59 | Miscellaneous | 625 |
| 10 | Authorizations | 816 | 60 | Modifications | 350 |
| 11 | Base Salary | 882 | 61 | No Conflicts | 953 |
| 12 | Benefits | 560 | 62 | No Defaults | 446 |
| 13 | Binding Effects | 807 | 63 | No Waivers | 604 |
| 14 | Books | 25 | 64 | Non-Disparagement | 464 |
| 15 | Brokers | 527 | 65 | Notices | 3313 |
| 16 | Capitalization | 565 | 66 | Organizations | 550 |
| 17 | Change In Control | 479 | 67 | Participations | 610 |
| 18 | Closings | 685 | 68 | Payments | 1157 |
| 19 | Compliance With Laws | 1335 | 69 | Positions | 256 |

Table A1: List of Provisions

| Class ID | Provision | Count | Class ID | Provision | Count |
|---|---|---|---|---|---|
| 20 | Confidentiality | 1328 | 70 | Powers | 153 |
| 21 | Consent To Jurisdiction | 243 | 71 | Publicity | 437 |
| 22 | Consents | 582 | 72 | Qualifications | 60 |
| 23 | Construction | 538 | 73 | Records | 408 |
| 24 | Cooperation | 618 | 74 | Releases | 772 |
| 25 | Costs | 205 | 75 | Remedies | 901 |
| 26 | Counterparts | 3346 | 76 | Representations | 755 |
| 27 | Death | 395 | 77 | Sales | 421 |
| 28 | Defined Terms | 482 | 78 | Sanctions | 156 |
| 29 | Definitions | 778 | 79 | Severability | 2552 |
| 30 | Disability | 354 | 80 | Solvency | 558 |
| 31 | Disclosures | 791 | 81 | Specific Performance | 417 |
| 32 | Duties | 473 | 82 | Submission To Jurisdiction | 309 |
| 33 | Effective Dates | 612 | 83 | Subsidiaries | 721 |
| 34 | Effectiveness | 306 | 84 | Successors | 631 |
| 35 | Employment | 397 | 85 | Survival | 1951 |
| 36 | Enforceability | 346 | 86 | Tax Withholdings | 512 |
| 37 | Enforcements | 289 | 87 | Taxes | 1488 |
| 38 | Entire Agreements | 3105 | 88 | Terminations | 1423 |
| 39 | Erisa | 439 | 89 | Terms | 1511 |
| 40 | Existence | 337 | 90 | Titles | 406 |
| 41 | Expenses | 1577 | 91 | Transactions With Affiliates | 475 |
| 42 | Fees | 788 | 92 | Use Of Proceeds | 1123 |
| 43 | Financial Statements | 859 | 93 | Vacations | 446 |
| 44 | Forfeitures | 398 | 94 | Venues | 170 |
| 45 | Further Assurances | 1316 | 95 | Vesting | 621 |
| 46 | General | 1329 | 96 | Waiver Of Jury Trials | 1025 |
| 47 | Governing Laws | 4243 | 97 | Waivers | 1199 |
| 48 | Headings | 933 | 98 | Warranties | 684 |
| 49 | Indemnifications | 1228 | 99 | Withholdings | 711 |

# Appendix B    Precision, Recall and F1-score for each Topic

Table B1: Precision, Recall and F1-score for each Topic

| Class ID | Precision | Recall | F1-score | Class ID | Precision | Recall | F1-score |
|---|---|---|---|---|---|---|---|
| 0 | 0.95 | 0.98 | 0.96 | 50 | 0.93 | 0.93 | 0.93 |
| 1 | 0.61 | 0.81 | 0.70 | 51 | 0.99 | 0.99 | 0.99 |
| 2 | 0.93 | 0.94 | 0.93 | 52 | 0.93 | 0.69 | 0.80 |
| 3 | 0.88 | 1.00 | 0.94 | 53 | 1.00 | 1.00 | 1.00 |
| 4 | 0.69 | 0.21 | 0.32 | 54 | 0.82 | 0.95 | 0.88 |
| 5 | 0.60 | 0.81 | 0.69 | 55 | 0.83 | 0.82 | 0.82 |
| 6 | 0.94 | 0.98 | 0.96 | 56 | 0.88 | 0.72 | 0.79 |
| 7 | 0.98 | 0.93 | 0.96 | 57 | 0.94 | 0.96 | 0.95 |
| 8 | 0.00 | 0.00 | 0.00 | 58 | 0.99 | 0.98 | 0.99 |
| 9 | 0.79 | 0.79 | 0.79 | 59 | 0.76 | 0.59 | 0.67 |
| 10 | 0.86 | 0.89 | 0.87 | 60 | 0.86 | 0.80 | 0.83 |
| 11 | 1.00 | 0.99 | 1.00 | 61 | 0.97 | 0.97 | 0.97 |
| 12 | 0.96 | 0.99 | 0.98 | 62 | 1.00 | 0.98 | 0.99 |
| 13 | 0.83 | 0.71 | 0.76 | 63 | 0.93 | 0.75 | 0.83 |
| 14 | 1.00 | 1.00 | 1.00 | 64 | 1.00 | 1.00 | 1.00 |
| 15 | 1.00 | 1.00 | 1.00 | 65 | 0.97 | 1.00 | 0.98 |
| 16 | 1.00 | 0.98 | 0.99 | 66 | 0.92 | 0.97 | 0.95 |
| 17 | 0.98 | 0.97 | 0.97 | 67 | 0.99 | 0.96 | 0.97 |
| 18 | 0.95 | 0.97 | 0.96 | 68 | 0.90 | 0.90 | 0.90 |
| 19 | 0.99 | 0.99 | 0.99 | 69 | 0.89 | 0.50 | 0.64 |
| 20 | 0.99 | 0.98 | 0.99 | 70 | 1.00 | 0.85 | 0.92 |
| 21 | 0.62 | 0.91 | 0.74 | 71 | 1.00 | 0.98 | 0.99 |
| 22 | 0.91 | 0.84 | 0.87 | 72 | 0.67 | 0.40 | 0.50 |
| 23 | 0.60 | 0.87 | 0.71 | 73 | 0.97 | 1.00 | 0.98 |
| 24 | 0.99 | 0.93 | 0.96 | 74 | 1.00 | 0.97 | 0.98 |
| 25 | 0.83 | 0.67 | 0.74 | 75 | 0.92 | 0.95 | 0.93 |
| 26 | 0.95 | 1.00 | 0.98 | 76 | 0.83 | 0.88 | 0.86 |
| 27 | 0.98 | 0.91 | 0.94 | 77 | 0.95 | 0.91 | 0.93 |
| 28 | 0.61 | 0.75 | 0.67 | 78 | 1.00 | 0.79 | 0.88 |
| 29 | 0.85 | 0.72 | 0.78 | 79 | 1.00 | 0.99 | 0.99 |
| 30 | 1.00 | 0.92 | 0.96 | 80 | 1.00 | 0.98 | 0.99 |
| 31 | 0.98 | 0.95 | 0.96 | 81 | 0.95 | 0.93 | 0.94 |
| 32 | 0.87 | 0.90 | 0.88 | 82 | 1.00 | 0.55 | 0.71 |
| 33 | 0.81 | 0.89 | 0.85 | 83 | 0.94 | 1.00 | 0.97 |
| 34 | 0.93 | 0.45 | 0.61 | 84 | 0.71 | 0.85 | 0.77 |
| 35 | 0.73 | 0.94 | 0.82 | 85 | 0.95 | 0.99 | 0.97 |
| 36 | 0.90 | 0.84 | 0.87 | 86 | 0.82 | 0.73 | 0.77 |
| 37 | 0.86 | 0.74 | 0.79 | 87 | 0.96 | 0.92 | 0.94 |
| 38 | 0.95 | 0.99 | 0.97 | 88 | 0.89 | 0.95 | 0.92 |
| 39 | 1.00 | 1.00 | 1.00 | 89 | 0.90 | 0.90 | 0.90 |
| 40 | 1.00 | 0.86 | 0.92 | 90 | 0.95 | 0.80 | 0.87 |
| 41 | 0.93 | 0.97 | 0.95 | 91 | 1.00 | 0.96 | 0.98 |
| 42 | 0.96 | 0.88 | 0.92 | 92 | 1.00 | 0.97 | 0.98 |
| 43 | 0.98 | 1.00 | 0.99 | 93 | 1.00 | 0.98 | 0.99 |
| 44 | 0.96 | 0.96 | 0.96 | 94 | 0.85 | 0.55 | 0.67 |
| 45 | 0.99 | 0.99 | 0.99 | 95 | 0.94 | 0.94 | 0.94 |

Table B1: Precision, Recall and F1-score for each Topic

| Class ID | Precision | Recall | F1-score | Class ID | Precision | Recall | F1-score |
|---|---|---|---|---|---|---|---|
| 46 | 0.84 | 0.91 | 0.87 | 96 | 1.00 | 1.00 | 1.00 |
| 47 | 0.93 | 0.98 | 0.95 | 97 | 0.86 | 0.96 | 0.91 |
| 48 | 1.00 | 0.95 | 0.97 | 98 | 0.95 | 0.75 | 0.84 |
| 49 | 0.98 | 0.97 | 0.98 | 99 | 0.80 | 0.88 | 0.84 |

# Funding

The authors declare no extra funding support for this study.

# Ethics Declarations

On behalf of all authors, the corresponding author states that there is no conflict of interest.

The authors have no relevant financial or non-financial interests to disclose.

The authors have no conflicts of interest to declare that are relevant to the content of this article.

All authors certify that they have no affiliations with or involvement in any organization or entity with any financial interest or non-financial interest in the subject matter or materials discussed in this manuscript.

The authors have no financial or proprietary interests in any material discussed in this article.